\documentclass{article}
\usepackage{PRIMEarxiv}
\usepackage[utf8]{inputenc}
\usepackage[T1]{fontenc}
\usepackage{lmodern}   % provides a bold typewriter face (T1/lmtt/bx)

\usepackage{amsmath,amsfonts,bm}

\def\eqref#1{equation~\ref{#1}}
\def\1{\bm{1}}

\DeclareMathAlphabet{\mathsfit}{\encodingdefault}{\sfdefault}{m}{sl}
\SetMathAlphabet{\mathsfit}{bold}{\encodingdefault}{\sfdefault}{bx}{n}

\usepackage[hyphens]{url}
\usepackage{booktabs}
\usepackage{amsfonts}
\usepackage{microtype}
\usepackage{graphicx}
\graphicspath{{figs/}}
\usepackage{multirow}
\usepackage{subcaption}
\usepackage{array}
\usepackage{makecell}
\usepackage[round]{natbib}
\usepackage[title]{appendix}
\usepackage[table]{xcolor}
\usepackage{tcolorbox}
\tcbuselibrary{skins,breakable}

\usepackage{float}
\usepackage{enumitem}
\usepackage{xspace}
\usepackage{pifont}
\usepackage{listings}
\definecolor{xlinkcolor}{cmyk}{1,0.6,0,0}
\usepackage[bookmarks=false,
     pdfnewwindow=true,
     colorlinks=true,
     linkcolor=xlinkcolor,
     citecolor=xlinkcolor,
     filecolor=xlinkcolor,
     urlcolor=xlinkcolor,
     breaklinks=true,
final=true]{hyperref}

\makeatletter
\g@addto@macro{\UrlBreaks}{\UrlOrds}
\makeatother

\newcommand{\modelname}{\textsc{ReToolSQL}\xspace}
\newcommand{\pp}{\%}          % percentage symbol in tables

\title{ReToolSQL: Agentic Reinforcement Learning \\ for Robust Text-to-SQL}

\author{
\textbf{Pratik Kakkar\thanks{Corresponding authors: \texttt{pratik.kakkar@jpmchase.com}, \texttt{chandra.dhir@jpmchase.com}},
Chandra Dhir$^*$,
Ravi Shankar,
Pareekshit Reddy Gaddam,
Anup Shirgaonkar} \\[0.5em]
\textbf{JPMorganChase}
}

\begin{document}
\maketitle
\let\thefootnote\relax\footnotetext{\textbf{Disclaimer:} \textit{This paper was prepared for informational purposes and is not a product of the Research Department of JPMorganChase. JPMorganChase makes no representation, warranty or undertaking whatsoever and disclaims all liability for the completeness, accuracy or reliability of the information contained herein. This document is not intended as investment research or investment advice, or a recommendation, offer or solicitation for the purchase or sale of any security, financial instrument, financial product or service, or to be used in any way for evaluating the merits of participating in any transaction, and shall not constitute a solicitation under any jurisdiction or to any person, if such solicitation under such jurisdiction or to such person would be unlawful.}}

% ── Abstract ──────────────────────────────────────────────────────
% !TEX root = ../main.tex
\begin{abstract}
Recent work has shown that reinforcement learning from execution feedback can substantially improve text-to-SQL performance, often enabling smaller models to match or exceed much larger systems.
However, most existing approaches treat SQL generation as a single-turn task, limiting the model's ability to recover from errors through iterative refinement.
We present \modelname, a two-stage training framework for text-to-SQL that combines (i)~a supervised warm-start on rejection-sampled reasoning traces with (ii)~agentic reinforcement fine-tuning~(RFT) over multi-turn tool-use trajectories.
The key insight is that the two stages act on complementary axes, the supervised fine-tuning~(SFT) on verified privileged-teacher traces expands the set of solvable questions (raising pass@$k$ coverage on the hardest cases), while RFT converts that expanded capability into higher single-pass accuracy by teaching the model when to verify, what evidence to retrieve, and how to repair faulty SQL from execution feedback.
Applied to Gemma~4 instruction-tuned (31B), RFT alone achieves 73.66\%~execution accuracy (EX) on the BIRD-SQL development benchmark (74.12\%~EX with self-consistency).
Initializing RFT from the SFT checkpoint (SFT$\to$RFT) yields our strongest model at 74.32\%~EX single-pass and 74.77\%~EX with self-consistency. At the time of writing, this ranked first on the BIRD single-model development-set leaderboard.
The approach uses composite rewards anchored on execution correctness, requires no human annotation beyond the benchmark itself, and operates within a single dense 31B model, showing that a properly designed SFT$\to$RFT pipeline over tool-use trajectories is a practical path toward robust enterprise-grade text-to-SQL.
\end{abstract}


% ── Main Sections ─────────────────────────────────────────────────
% !TEX root = ../main.tex
\section{Introduction}
\label{sec:intro}

Text-to-SQL is a promising interface for natural-language data access, but robust deployment remains difficult in realistic settings where schemas are large, values are noisy, and correctness depends on execution rather than surface-form similarity.
Benchmarks such as BIRD~\citep{li2023bird} highlight this challenge by evaluating systems through execution accuracy~(EX) on complex cross-domain databases, exposing persistent failures in schema linking, value grounding, and numerical or temporal reasoning.
Although recent instruction-tuned large language models can generate correct SQL for many questions, they remain brittle as a single incorrect join, literal, or aggregation often causes failure, and single-pass generation offers no mechanism for recovery.

Recent work shows that reinforcement learning from execution feedback can substantially improve text-to-SQL performance, often allowing relatively smaller models to rival much larger systems~\citep{yao2026arctic, shao2024deepseekmath}.
Yet most existing methods still treat SQL generation as a single-turn act: the model proposes one query, receives a terminal reward, and is not explicitly trained to use execution feedback as part of an iterative reasoning process.
At the same time, many of the strongest systems rely on increasingly heavyweight pipelines with multiple generators, selectors, or orchestration stages~\citep{pourreza2025chase, liu2026xiyan, talaei2024chess}, improving accuracy at the cost of inference-time complexity.
This motivates a central question: \emph{can a single dense model learn to propose, verify, and repair its own SQL through tool interaction while retaining a simple serving architecture?}

\begin{figure}[t]
\centering
\includegraphics[width=\linewidth]{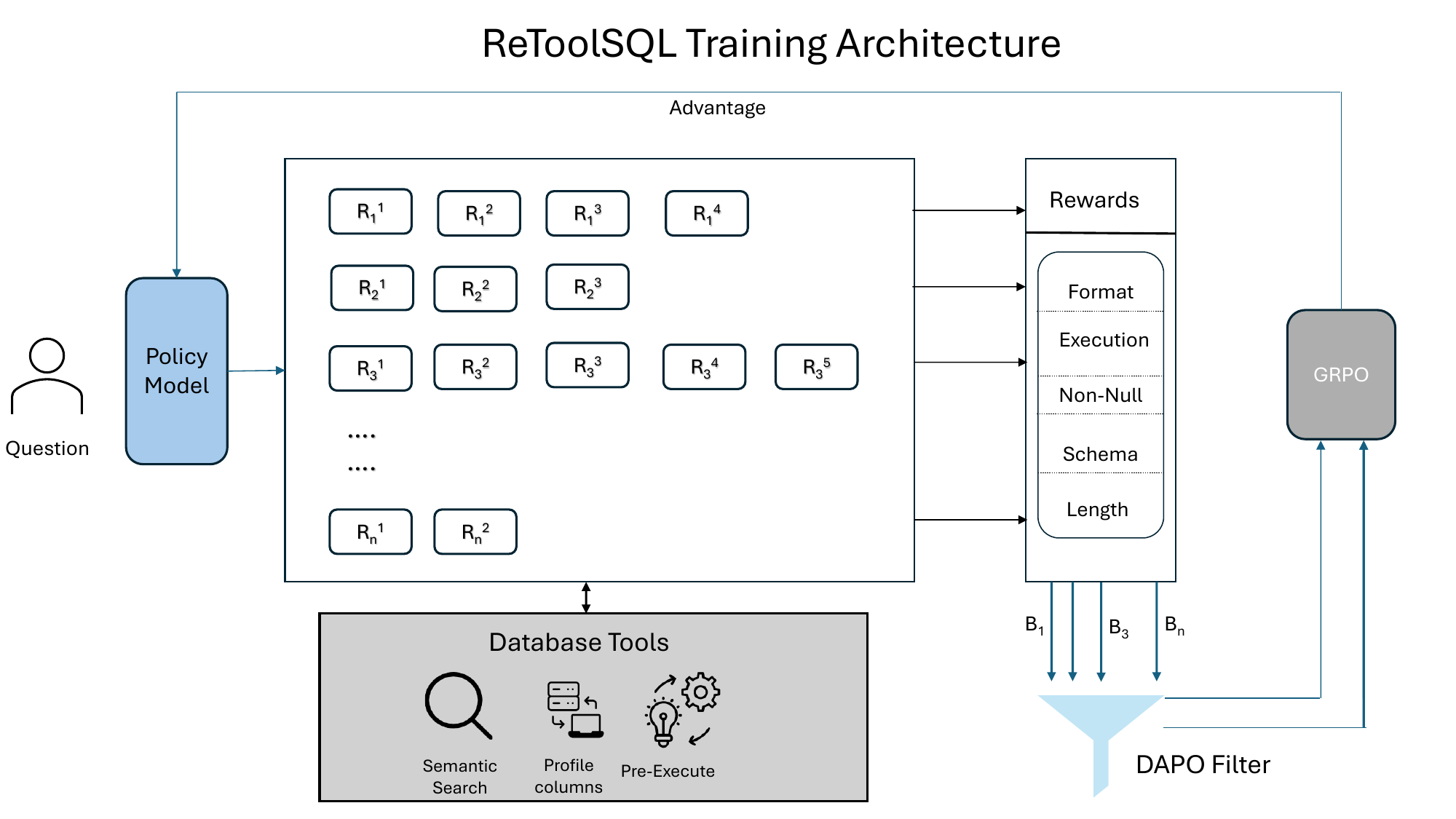}
\caption{High-level overview of \modelname. A natural-language question and enriched schema are processed by the Gemma~4 31B policy, which interacts with database tools across multiple turns to produce verified SQL. In the training loop, $B_1, B_2, \ldots$ denote the prompt groups sampled per training step (each containing $G{=}16$ multi-turn trajectories for a single question), and $R_1, R_2, \ldots$ the trajectory-level composite rewards $R=\sum_k w_k R_k$ (Section~\ref{sec:reward}) that yield the group-relative advantages driving the GRPO update. Groups with zero reward variance ($\text{std}=0$) are discarded by dynamic sampling (Section~\ref{sec:dynamic-sampling}), as they contribute no gradient signal.}
\label{fig:overview}
\end{figure}

We address this question with \modelname\footnote{\textbf{Re}inforcement-learned \textbf{Tool}-use for \textbf{SQL}.}, a text-to-SQL system trained with agentic reinforcement learning over multi-turn tool-use trajectories (Figure~\ref{fig:overview}).
Our approach adapts the tool-integrated RL recipe of ReTool~\citep{feng2026retool}, which teaches an LLM \emph{when} and \emph{how} to invoke a code interpreter for mathematical reasoning, to the task of natural-language-to-SQL generation. In place of the code interpreter, we equip the policy with a suite of read-only database tools (a sandboxed SQL executor, a column profiler, and a BM25 value search) and add schema-linking reward terms, so that tool use is grounded in real database evidence specific to text-to-SQL.
Instead of treating database execution as a post-hoc filter, \modelname places these read-only tools directly inside the training environment and learns from their execution feedback: the policy can execute candidate SQL, inspect runtime feedback, profile uncertain columns, and search stored values before committing to a final answer.
Training uses a GRPO-style objective~\citep{shao2024deepseekmath} over complete propose--verify--repair trajectories with a composite reward anchored on execution correctness and schema-linking signals.
As a result, the policy learns not only to produce SQL, but also \emph{when} to verify, \emph{what} evidence to retrieve, and \emph{how} to revise a faulty draft in response to the database feedback.

Applied to Gemma~4 instruction-tuned (31B), agentic RL alone achieves 73.66\% execution accuracy on the BIRD-SQL development set (74.12\% with self-consistency).
Initializing RL from a supervised fine-tuning (SFT) checkpoint trained on rejection-sampled reasoning traces (a two-stage SFT$\to$RFT pipeline) yields our strongest model at 74.32\% single-pass and 74.77\% with self-consistency, improving over both individual stages. The SFT warm-start expands pass@16 coverage on hard questions, and agentic RL converts that expanded coverage into higher single-pass accuracy.
Together, these results show that a carefully designed SFT$\to$RFT pipeline over tool-use trajectories is a practical path toward robust, enterprise-grade text-to-SQL within a single dense model.

\paragraph{Contributions.}
\begin{enumerate}[nosep,leftmargin=*]
  \item We formulate text-to-SQL as a tool-augmented sequential decision problem and introduce a verification-first reinforcement learning setup with execution-grounded tool use, DAPO-style dynamic sampling, and a composite reward (Section~\ref{sec:method}).
  \item We propose a two-stage SFT$\to$RFT pipeline: a supervised warm-start on execution-verified, rejection-sampled privileged-teacher traces followed by agentic RL, and show that the SFT expands pass@$k$ coverage on hard questions while RL improves the single-pass accuracy (Section~\ref{sec:results}).
  \item We show that a single dense 31B model ranks \#1 on the BIRD-SQL single-model dev leaderboard, competitive with substantially more complex pipeline based systems, and provide extensive ablations over model scale, entropy preservation, and inference-time scaling dynamics (Sections~\ref{sec:results},~\ref{sec:analysis}).
\end{enumerate}
% !TEX root = ../main.tex
\section{Related Work}
\label{sec:related}

Our approach sits at the intersection of four lines of work: LLM-based text-to-SQL, pipeline-based test-time scaling, reinforcement learning with verifiable rewards~(RLVR), and agentic tool use for execution-grounded generation.

% ──────────────────────────────────────────────────────────────────
\subsection{LLM-Based Text-to-SQL}

Text-to-SQL has evolved from grammar constrained sequence-to-sequence models on Spider~\citep{yu2018spider} to instruction-tuned LLMs evaluated on BIRD~\citep{li2023bird}.
BIRD is particularly challenging because it spans 12,751 question--SQL pairs over 95 large, noisy databases across more than 37 domains and evaluates systems by execution accuracy, set equality between predicted and gold query outputs, rather than by string or tree similarity.
This shifts the problem from surface-form matching toward schema linking, value grounding, and numeric or temporal fidelity.
Prompt-based decomposition methods such as DIN-SQL~\citep{pourreza2023din} address this difficulty by separating linking from generation, while representation-focused work such as M-Schema serialization~\citep{liu2026xiyan} improves schema encoding by exposing types, value exemplars, and foreign-key structure.
Our context builder follows this latter perspective by injecting column meanings, cardinalities, value ranges, and representative values so that linking decisions are grounded in evidence rather than names alone.

% ──────────────────────────────────────────────────────────────────
\subsection{Pipelines, Candidate Selection, and Test-Time Scaling}

Recent BIRD leaders rely heavily on orchestrated pipelines that generate many candidates and then select among them. CHESS~\citep{talaei2024chess} deploys four specialized agents; information retrieval, schema selection, candidate generation, and unit testing achieving 71.1\%~EX on BIRD test with open-source models.
CHASE-SQL~\citep{pourreza2025chase} combines divide-and-conquer prompting, execution-plan reasoning, and instance-aware few-shot synthesis with a learned pairwise selector (73.0\%~EX). XiYan-SQL~\citep{liu2026xiyan} couples M-Schema with a multi-generator ensemble and candidate selection (75.63\%~EX), while Agentar-Scale-SQL~\citep{wang2025agentar} further expands this paradigm through iterative refinement and tournament-style selection, reaching 81.67\%~EX on BIRD test.
AgentNLQ~\citep{bogdanov2026agentnlq} likewise employs a multi-agent orchestrator that plans, reflects, and self-corrects over a semantically enriched schema augmented with user-provided business rules, reporting 78.1\%~semantic accuracy on BIRD.
These results demonstrate that test-time compute is a strong lever for performance, but they also increase serving cost and system complexity.
By contrast, our approach internalizes proposal, verification, and repair within a single dense policy, reducing test-time scaling to execution-based self-consistency~\citep{wang2022self} instead of a learned selector over heterogeneous generators.

% ──────────────────────────────────────────────────────────────────
\subsection{Reinforcement Learning with Verifiable Rewards}

RLVR has emerged as a practical framework for eliciting reasoning in LLMs.
GRPO~\citep{shao2024deepseekmath} removes the value network of classical actor-critic RL by estimating advantages from reward statistics over groups of sampled completions. DeepSeek-R1~\citep{guo2025deepseekr1} shows that verifiable terminal rewards alone can induce long-horizon reasoning. DAPO~\citep{yu2026dapo} introduces stabilizers such as asymmetric clipping, dynamic sampling to discard homogeneous prompt groups, and soft overlength penalties to prevent entropy collapse.
Closest to our setting, Arctic-Text2SQL-R1~\citep{yao2026arctic} trains a 7B model with GRPO under a binary execution-match reward and outperforms many much larger systems on BIRD.

We build on this line of work in three ways:
(i)~we apply DAPO-style dynamic sampling to filter prompt groups that provide no learning signal,
(ii)~we use a composite, binary-heavy reward that retains schema-linking signal when execution rewards collapse to uniform values within a group, and
(iii)~we collect reward over multi-turn agentic trajectories rather than single-shot generations, so that the model is optimized for the entire propose--verify--repair cycle.

% ──────────────────────────────────────────────────────────────────
\subsection{Agentic Tool Use and Execution Grounding}

ReAct~\citep{yao2023react} established the value of interleaving reasoning with tool actions, and subsequent work on multi-turn tool agents~\citep{chen2023fireact} showed that fine-tuning on agentic trajectories can substantially improve tool reliability. Most closely related, ReTool~\citep{feng2026retool} trains an LLM with outcome-driven RL to interleave real-time code execution with reasoning and to learn when and how to invoke tools from execution feedback, pairing a cold-start supervised stage with subsequent RL. We adopt the same two-stage recipe, a supervised warm-start followed by reinforcement learning, but target database tools and execution-grounded SQL instead of a code interpreter for mathematical reasoning. In the database setting, the relevant tools include a sandboxed SQL executor, a column profiler, and a lexical value-search mechanism. Prior text-to-SQL systems typically use execution only as a post-hoc filter or for offline data curation. We instead place tools directly inside the RL environment and reward their effective use during training. This enables the model to learn a verification-first policy: it drafts a query, executes it, and inspects errors or output shape. When ambiguities remain, it profiles columns to resolve type or scale issues and searches stored values with BM25~\citep{robertson2009bm25} to repair literal mismatches. All of this happens before the model produces the final SQL.

The key distinction of our method compared to pipeline systems is that tool proficiency emerges from gradient updates on complete trajectories, as opposed to being a property of the prompt or an external orchestration layer. \modelname thus combines agentic tool use during RL training, GRPO with dynamic sampling, a composite schema-aware reward, and execution-based self-consistency decoding within a single dense 31B model.
% !TEX root = ../main.tex
\section{Methodology}
\label{sec:method}

We formulate text-to-SQL as a tool-augmented sequential decision problem solved by reinforcement fine-tuning, and we decode with a training-free execution-consensus rule. The approach is model-agnostic; in our experiments the policy is instantiated with a single dense, instruction-tuned model with 31B parameters (Gemma~4 31B instruct), served with a vLLM serving engine. No external retriever, or auxiliary selection model is required.
Figure~\ref{fig:overview} gives the end-to-end picture where a question is augmented with an enriched schema, the policy generates multi-turn agentic rollouts that interact with database tools, each terminal SQL query is scored by the composite reward, and group-relative advantages drive the GRPO update with DAPO-style dynamic sampling.

% ──────────────────────────────────────────────────────────────────
\subsection{Problem Formulation}
\label{sec:formulation}

We model an episode as a Partially Observable Markov Decision Process (POMDP) instantiated per example by the tuple $(\mathcal{Q}, \mathcal{S}, \mathcal{T})$, where $\mathcal{Q}$ is the natural-language question concatenated with optional evidence (a ``hint''), $\mathcal{S}$ is the schema of the target database $\mathcal{D}$, and $\mathcal{T}$ is a fixed set of read-only tools.
The policy $\pi_\theta$ is a single autoregressive language model.
At turn $t$ it observes a context $c_t$ and emits an action $a_t$ that is either a tool call $a_t^{\text{tool}}$ or the terminal answer $a_t^{\text{sql}}$ (the final SQL query).
The environment returns an observation $o_t$ (tool output or error), and the context is extended autoregressively:
\begin{equation}
\label{eq:context}
  c_{t+1} = [c_t; a_t; o_t].
\end{equation}

A trajectory $\tau = (c_0, a_0, o_0, \ldots, a_T)$ terminates when the model emits \texttt{<final\_answer>} or the per-episode tool budget ($T_{\max} = 8$ turns) is exhausted. The objective is execution correctness of $a_T^{\text{sql}}$ against $\mathcal{D}$, not token-level imitation of a gold reasoning trace or reference SQL.

% ──────────────────────────────────────────────────────────────────
\subsection{Schema-Grounded Observation Space}
\label{sec:schema}

The single largest determinant of correct SQL is the quality of the schema representation. We adopt an M-Schema~\citep{liu2026xiyan} encoding which is a semi-structured, value-enriched serialization that is consistently stronger than raw data definition language (DDL) or flat column lists because it exposes types, foreign keys, and data evidence in a compact, regular layout the model can attend to.

Our builder emits, per database, a \texttt{db\_id} header and a \texttt{[Schema]} block. Each table is introduced by \texttt{\# Table:} with its comment and row/column counts, followed by a bracketed list of columns (Figure~\ref{lst:schema}).
Every column line carries its name and type, an inline natural-language meaning drawn from a per-database dictionary, key/nullability flags, a compact statistics group (Nulls, Non-null, Distinct, Min/Avg/Max), the most frequent categorical \emph{Top} values, and a few literal \emph{Examples}; a trailing \texttt{[Foreign keys]} block lists join paths.
This representation enables evidence-based schema linking so that the policy selects columns and constructs predicates from observed value distributions instead of from column-name similarity.

% \begin{lstlisting}[style=mschema,caption={M-Schema encoding example (abbreviated).},label={lst:schema}]
% `california_schools`
% [Schema]
% # Table: frpm, Free/Reduced-Price Meal eligibility | Rows: 9986, Columns: 28
% [
% (CDSCode:TEXT, County-District-School code, Primary Key, Not Null,
%    Stats: Distinct: 9986, Examples: ['01100170109835','01100170112607']),
% (County Name:TEXT, county of the school, Nullable,
%    Stats: Nulls: 0, Distinct: 58, Top: ['Los Angeles','Orange','San Diego']),
% (Free Meal Count (K-12):REAL, K-12 free meal eligible, Nullable,
%    Stats: Nulls: 1043, Min: 0.0, Avg: 287.4, Max: 5333.0)
% ]
% # Table: schools | Rows: 17686, Columns: 49
% [
% (CDSCode:TEXT, Primary Key, Not Null),
% (Charter:INTEGER, 1 = charter, 0 = non-charter, Nullable, Distinct: 2)
% ]
% [Foreign keys]
% frpm.CDSCode=schools.CDSCode
% \end{lstlisting}

\begin{figure}[t]
\definecolor{schemaframe}{HTML}{3B6E8F}
\definecolor{tablehdr}{HTML}{2B579A}
\definecolor{colname}{HTML}{1A6B3F}
\definecolor{coltype}{HTML}{8B4513}
\definecolor{statsclr}{HTML}{555555}
\definecolor{fkcolor}{HTML}{7B2D8B}
\begin{tcolorbox}[
  enhanced,
  title={\small\texttt{california\_schools} --- M-Schema encoding (abbreviated)},
  fonttitle=\sffamily\bfseries,
  coltitle=white,
  colbacktitle=schemaframe!85!black,
  colback=gray!2,
  colframe=schemaframe!50!black,
  boxrule=0.5pt,
  arc=2pt,
  left=6pt, right=6pt, top=4pt, bottom=4pt,
]

{\sffamily\bfseries\color{tablehdr}\# Table: frpm}\enspace{\small\color{tablehdr!70!black}Free/Reduced-Price Meal eligibility\enspace|\enspace Rows: 9{,}986\enspace|\enspace Columns: 28}

\smallskip
\begin{tabular}{@{}p{0.97\linewidth}@{}}
{\small\ttfamily\color{colname}CDSCode}{\small\color{coltype}:TEXT}\enspace{\scriptsize PK, Not Null}\\[-1pt]
{\scriptsize\color{statsclr}\quad Stats: Distinct: 9986 \quad Examples: \texttt{['01100170109835','01100170112607']}}\\[3pt]
{\small\ttfamily\color{colname}County Name}{\small\color{coltype}:TEXT}\enspace{\scriptsize county of the school, Nullable}\\[-1pt]
{\scriptsize\color{statsclr}\quad Stats: Nulls: 0, Distinct: 58 \quad Top: \texttt{['Los Angeles','Orange','San Diego']}}\\[3pt]
{\small\ttfamily\color{colname}Free Meal Count (K-12)}{\small\color{coltype}:REAL}\enspace{\scriptsize K-12 free meal eligible, Nullable}\\[-1pt]
{\scriptsize\color{statsclr}\quad Stats: Nulls: 1043, Min: 0.0, Avg: 287.4, Max: 5333.0}\\
\end{tabular}

\tcbline

{\sffamily\bfseries\color{tablehdr}\# Table: schools}\enspace{\small\color{tablehdr!70!black}Rows: 17{,}686\enspace|\enspace Columns: 49}

\smallskip
\begin{tabular}{@{}p{0.97\linewidth}@{}}
{\small\ttfamily\color{colname}CDSCode}{\small\color{coltype}:TEXT}\enspace{\scriptsize PK, Not Null}\\[3pt]
{\small\ttfamily\color{colname}Charter}{\small\color{coltype}:INTEGER}\enspace{\scriptsize 1 = charter, 0 = non-charter, Nullable, Distinct: 2}\\
\end{tabular}

\tcbline

{\sffamily\bfseries\color{fkcolor}[Foreign Keys]}\par
\smallskip
{\small\ttfamily\color{fkcolor} frpm.CDSCode = schools.CDSCode}

\end{tcolorbox}
\captionof{figure}{M-Schema encoding example (abbreviated). Each column carries its type, semantics, nullability, summary statistics, and representative values, enabling evidence-based schema linking.}
\label{lst:schema}
\end{figure}

Few-shot exemplars are retrieved with BM25~\citep{robertson2009bm25} over the training pool, excluding any example sharing the target \texttt{db\_id} to prevent leakage.
During training we cap the prompt at 15{,}500 tokens and the completion at 8{,}000 tokens. Prompts which are longer than this are filtered instead of truncated so that no schema evidence is silently dropped.

% ──────────────────────────────────────────────────────────────────
\subsection{Action Space: A Verification-First Tool Suite}
\label{sec:tools}

The policy interacts with the environment $\mathcal{T}$ through three deterministic, read-only tools (full signatures and parameters are discussed in Appendix~\ref{app:tools}):

\begin{description}[itemsep=6pt,leftmargin=1em]
  \item[\textbf{\texttt{bm25\_search\_sqlite}}:] A hybrid lexical search over stored cell values that combines BM25 ranking with a regex/substring fallback. It grounds string and category literals. For example, mapping a paraphrased entity to its exact stored spelling so that \texttt{WHERE} predicates match real values.
  \item[\textbf{\texttt{sqlite\_query}}:] A sandboxed executor restricted to \texttt{SELECT}/\texttt{WITH} statements, with a busy timeout and a returned-row cap. It reports either the result columns and rows or a structured execution error, enabling execution-guided self-repair.
  \item[\textbf{\texttt{sqlite\_peek}}:] A column profiler returning declared type, null/distinct counts, min/max, representative samples, character-class summaries, and value prefixes. It resolves the ambiguities that most often break BIRD queries, such as percent-vs-fraction scales, stored date formats, and join-key type mismatches.
\end{description}

We impose a strict turn protocol that doubles as a learnable structure.
Every assistant turn contains a bounded \texttt{<scratch\_pad>} (intent, expected output columns, and the candidate SQL) followed by exactly one of a single native tool call or a \texttt{<final\_answer>} block.
Immediately before finalizing, the policy must emit explicit \texttt{<relevant\_tables>} and \texttt{<relevant\_columns>} annotations covering every table and column used by the final SQL. These annotations materialize the schema-linking decision and are consumed directly by the linking rewards as discussed in Section~\ref{sec:reward}.

The prescribed workflow drafts SQL from schema and hint, executes it with \texttt{sqlite\_query}, and invokes \texttt{sqlite\_peek} or \texttt{bm25\_search\_sqlite} only to repair a detected fault. Appendix~\ref{app:example} walks through a complete multi-turn trajectory.
Tools are framed as instruments of verification and grounding, not as a substitute for the model's own SQL reasoning.

% ──────────────────────────────────────────────────────────────────
\subsection{Policy Optimization: GRPO with Asymmetric Clipping}
\label{sec:grpo}

We optimize $\pi_\theta$ with Group Relative Policy Optimization (GRPO;~\citep{shao2024deepseekmath}) using the DAPO loss variant~\citep{yu2026dapo}.
For each question $q$ we sample a group of $G = 16$ complete trajectories $\{o_i\}_{i=1}^G$ from the behavior policy $\pi_{\theta_{\text{old}}}$ and score each with the composite reward $R_i$ of Section~\ref{sec:reward}.
Advantages are computed by group-relative normalization, which removes the need for a learned value function:
\begin{equation}
\label{eq:advantage}
  \hat{A}_i = \frac{R_i - \mu_G}{\sigma_G + \epsilon}\,,
\end{equation}
where $\mu_G$ and $\sigma_G$ are the mean and standard deviation of rewards within the group, and $\epsilon = 10^{-8}$ prevents division by zero.

The clipped surrogate objective uses asymmetric bounds following DAPO's clip higher prescription, augmented with a scheduled KL penalty toward a fixed reference policy:
\begin{equation}
\label{eq:grpo-loss}
  \mathcal{L}(\theta) = -\mathbb{E}_{q, i}\left[\min\!\left(\rho_i(\theta)\,\hat{A}_i,\; \text{clip}\bigl(\rho_i(\theta),\, 1{-}\varepsilon_{\text{lo}},\, 1{+}\varepsilon_{\text{hi}}\bigr)\,\hat{A}_i\right) - \beta_t\, D_{\mathrm{KL}}\!\bigl(\pi_\theta \,\|\, \pi_{\text{ref}}\bigr)\right],
\end{equation}
where $\rho_i(\theta) = \pi_\theta(o_i \mid q) / \pi_{\theta_{\text{old}}}(o_i \mid q)$ is the token-level importance ratio, $(\varepsilon_{\text{lo}}, \varepsilon_{\text{hi}}) = (0.20, 0.28)$, $\pi_{\text{ref}}$ is the frozen initial policy, and $\beta_t$ is the KL coefficient at training step $t$ (annealed per the schedule below).
The asymmetric upper clip allows the policy to increase probability on rare high-reward trajectories more aggressively than it decreases probability on failures, preserving upward exploration and preventing premature entropy collapse.
We use a single optimizer pass per rollout buffer ($\mu = 1$).
Instead of disabling KL regularization outright, we apply a decaying KL-penalty schedule ($\beta\!:\,0.005 \to 0.001 \to 0$) that preserves policy entropy and thus exploration capacity early in training before annealing to zero, after which reward shaping and asymmetric clipping alone regularize the policy. We compare this schedule against a constant $\beta = 0$ in Section~\ref{sec:beta}.

% ──────────────────────────────────────────────────────────────────
\subsection{Dynamic Sampling}
\label{sec:dynamic-sampling}

A persistent failure mode of group-relative RL is the \emph{degenerate group} where all $G$ trajectories for a prompt receive identical execution-match reward, then $\sigma_G = 0$, every $\hat{A}_i = 0$, and the prompt contributes no gradient while still costing a full rollout and log-probability pass.
We adopt DAPO's oversample-and-replace strategy~\citep{yu2026dapo}:
\begin{enumerate}[nosep,leftmargin=*]
  \item For each training step requiring $B$ prompt groups per rank, we draw $B \times K$ candidate prompts from a shuffled backup pool (oversample factor $K = 12$),
  \item We generate $G = 16$ completions for each candidate in a single vLLM call,
  \item We compute \texttt{result\_reward} (EX) for all completions and check heterogeneity, i.e.,  a group is retained if $\text{std}(R_{\text{group}}) \geq 10^{-6}$, and 
  \item We select up to $B$ heterogeneous groups. Any unfilled slots are zero-masked (contributing no gradient).
\end{enumerate}

Candidate rewards are evaluated before the expensive trainer-side log-probability forward pass, so compute is spent only on informative groups.

% ──────────────────────────────────────────────────────────────────
\subsection{Tool-Observation Loss Masking}
\label{sec:masking}

A multi-turn trajectory interleaves policy-generated tokens with environment injected observations executor result sets, column profiles, and BM25 hits all of which contain raw database content. Optimizing naively over the full completion would train the policy to predict returned rows, simultaneously rewarding memorization of specific database contents and corrupting the policy gradient with tokens the model never chose.

We, therefore, maintain a token-level tool mask $m^{\text{tool}}_j$ that is $0$ over every observation span and $1$ over policy-emitted tokens, and optimize under the effective mask:
\begin{equation}
\label{eq:mask}
  m_j = m^{\text{comp}}_j \odot m^{\text{tool}}_j\,,
\end{equation}
where $m^{\text{comp}}$ is the standard padding or truncation mask and $\odot$ denotes elementwise product.
Both the importance ratio $\rho_i(\theta)$ and the loss are computed only at positions where $m_j = 1$.
Crucially, observation tokens remain in the attention context (they must stay visible so later turns can condition on retrieved evidence), but they contribute to neither the ratio nor the objective~\citep{chen2023fireact}.
Gradients thus flow only through the model's own reasoning, tool invocations, and final SQL, so the policy learns when and how to inspect a database rather than to reproduce its contents, a property that also guards against train/test schema leakage.

% ──────────────────────────────────────────────────────────────────
\subsection{Length Regularization}
\label{sec:length}

To discourage truncation noise without hard-clipping useful reasoning, we add DAPO's Soft Overlong Punishment~\citep{yu2026dapo} as a continuous reward term $R_{\text{len}}$:
\begin{equation}
\label{eq:length-penalty}
  R_{\text{len}} = \begin{cases}
    0 & \text{if } L \leq L_{\text{target}}, \\
    -\alpha \cdot \frac{L - L_{\text{target}}}{L_{\text{max}} - L_{\text{target}}} & \text{if } L_{\text{target}} < L \leq L_{\text{max}}, \\
    -\alpha & \text{if } L > L_{\text{max}},
  \end{cases}
\end{equation}
where $L$ is the completion length in tokens, $L_{\text{target}} = 6{,}000$, $L_{\text{max}} = 8{,}000$, and $\alpha = 0.5$ (measured in reward units).

% ──────────────────────────────────────────────────────────────────
\subsection{Composite Reward Specification}
\label{sec:reward}

Optimizing the binary BIRD~EX signal alone yields sparse, frequently-degenerate groups.
We therefore decompose the reward into seven verifiable components, six non-negative terms plus the length penalty, combined as a weighted sum $R = \sum_{k=0}^{6} w_k R_k$ (Table~\ref{tab:reward}).
The reward is deliberately binary-heavy and result dominated with the execution-result term carrying the largest weight so the policy is pulled toward execution-equivalent answers, while the auxiliary terms keep advantages non-flat in groups where the result term collapses to all-zero or all-one.
The weights are set heuristically rather than tuned: the execution term is assigned the dominant weight to anchor optimization on correctness, while the auxiliary terms take small values chosen only to keep group advantages non-degenerate. We did not perform a weight search, and performance is not sensitive to the precise auxiliary values as long as the execution term dominates.
\begin{table}[ht]
\centering\small
\caption{Composite reward components and their weights $w_k$. The execution-result term $R_{\text{exec}}$ dominates, while linking terms supply dense schema-grounding signal. The final column lists the \emph{already-weighted} contribution $w_k R_k$ of each term (not the raw signal $R_k$); summing the positive upper bounds gives a maximum attainable reward of $3.8$.}
\label{tab:reward}
\begin{tabular}{lllcc}
\toprule
\textbf{Component} & \textbf{Type} & \textbf{Signal} & \textbf{$w_k$} & \textbf{Weighted range ($w_k R_k$)} \\
\midrule
$R_{\text{format}}$  & binary      & Correct output template compliance & 0.2 & $\{0, 0.2\}$ \\
$R_{\text{syntax}}$  & binary      & Predicted SQL parses and executes  & 0.5 & $\{0, 0.5\}$ \\
$R_{\text{exec}}$    & binary      & BIRD EX: row-set equality with gold & 2.0 & $\{0, 2.0\}$ \\
$R_{\text{tables}}$  & binary      & Predicted table set $\supseteq$ gold table set & 0.5 & $\{0, 0.5\}$ \\
$R_{\text{columns}}$ & continuous  & Jaccard(pred columns, gold columns)  & 0.5 & $[0, 0.5]$ \\
$R_{\text{nonempty}}$& binary      & Executes and returns $\geq 1$ non-null cell & 0.1 & $\{0, 0.1\}$ \\
$R_{\text{len}}$     & continuous  & DAPO soft overlong penalty          & 0.1 & $[-0.1, 0]$ \\
\bottomrule
\end{tabular}
\end{table}

Result equivalence uses the official BIRD semantics (unordered set comparison on raw rows) with a per-query execution timeout of 30\,seconds to avoid penalizing slow gold queries.

% ──────────────────────────────────────────────────────────────────
\subsection{Inference-Time Scaling via Execution Consensus}
\label{sec:consensus}

At test time we apply a training-free execution consensus that realizes test-time scaling without any learned selector (Appendix~\ref{app:self-consistency} details the full procedure).
For a question we sample $N = 16$ candidate trajectories at a moderate temperature ($T = 1.2$, which preserves vote diversity better than near-greedy decoding).
Each candidate SQL is normalized (comment stripping, sanitization, and a \texttt{SELECT}-only guard) and executed against $\mathcal{D}$ under the same safety constraints used in training.
Candidates that error or return an empty result set are discarded. We return the representative SQL of the largest cluster, breaking ties by earliest sample index and then by shorter SQL:
\begin{equation}
\label{eq:consensus}
  \hat{y} = \arg\max_{c \in \mathcal{C}} |\{i : \text{exec}(y_i) = \text{exec}(c)\}|\,,
\end{equation}
where $\mathcal{C}$ is the set of distinct execution results and $y_i$ is the $i$-th candidate SQL.
Because semantically distinct hallucinations tend to scatter into many small separate clusters while correct programs agree on one execution result, consensus concentrates probability mass on the robust answer and consistently improves over greedy top-1 decoding.

% ──────────────────────────────────────────────────────────────────
\subsection{SFT Warm-Start via Privileged Reasoning Traces}
\label{sec:sft}

A key finding of this work is that reinforcement fine-tuning is substantially more effective when initialized from a supervised warm-start instead of being applied directly to the instruction-tuned base model~\citep{yao2026arctic}.
The rationale is when the base policy assigns near-zero probability to correct trajectories for hard examples, RL training groups become degenerate (all-wrong), providing no gradient signal (Section~\ref{sec:dynamic-sampling}).
The SFT stage addresses this by expanding the model's coverage on hard examples before RL begins.
The key challenge is generating high-quality reasoning traces for the hardest examples, precisely those the base model cannot solve.

\paragraph{Trace generation.}
We mine the training set with pass@16 decoding to identify an \emph{all-wrong band}. These are the questions for which every sampled trajectory fails execution match.
For these hard examples, we generate \emph{teacher traces} by providing the model with privileged access to the gold reference SQL as contextual guidance, prompting it to produce a step-by-step reasoning chain that arrives at the correct answer.
Traces are verified by executing the final predicted SQL against the database; only traces that pass execution match are retained.
To maximize coverage, we run both greedy decoding and pass@8 sampling under the privileged prompt, recovering traces for examples that greedy decoding alone misses.

For easier examples (those the model already solves), we collect the model's own correct reasoning trajectories generated without any privileged information, ensuring no distribution shift on questions within the model's current capability.

\paragraph{Dataset assembly and training.}
The final SFT dataset merges verified teacher traces (hard band) with self-traces (easy/medium band), filtered to remove any instance where gold SQL or privileged information leaks into the reasoning chain.
We fine-tune with a standard causal language modeling objective, masking the prompt tokens so loss is computed only over the reasoning and SQL tokens.
The resulting SFT checkpoint serves as the initialization for GRPO training described in Section~\ref{sec:grpo}, yielding a two-stage SFT$\to$RFT pipeline.
As we show in Section~\ref{sec:results}, by ensuring non-degenerate reward variance on hard examples from the first RL step, it enables the policy gradient to improve on precisely the questions that direct RFT from the base model cannot address.
% !TEX root = ../main.tex
\section{Experimental Setup}
\label{sec:setup}

% ──────────────────────────────────────────────────────────────────
\subsection{Dataset}

We evaluate \modelname on the BIRD-SQL benchmark~\citep{li2023bird}, a challenging text-to-SQL testbed spanning 11 large databases with complex schemas, noisy values, and domain-specific terminology.
For RL training we use 6{,}601 question from the BIRD training split, spanning 69 databases that are disjoint from the development databases used for evaluation.
Unless otherwise mentioned, we report results on the BIRD full development set (1,534 questions across 11 databases), which provides a broad measure of cross-domain generalization.
The development set is annotated with three difficulty levels: Simple (860 questions), Moderate (443 questions), and Challenging (231 questions).

% ──────────────────────────────────────────────────────────────────
\subsection{Evaluation Metrics}

\paragraph{Execution Accuracy (EX).}
The predicted SQL and gold SQL are executed against the same database, and the resulting outputs are converted to unordered sets of tuples before comparison.
A prediction is correct if and only if the two sets are identical.
This protocol makes EX order-invariant and consistent with the official BIRD evaluation procedure.
We use a 30-second per-query execution timeout and 8 parallel workers. Following BIRD conventions, we report accuracy separately for Simple, Moderate, and Challenging questions.
Gains are often non-uniform across difficulty tiers, and per-tier analysis reveals where the improvements came from.
%whether improvements come from breadth (more simple questions correct) or depth (harder questions solved).

% ──────────────────────────────────────────────────────────────────
\subsection{Models and Baselines}

% \paragraph{Policy Model.}
Our primary experiments use Gemma~4 31B Instruct~\citep{gemma2026}, a dense decoder-only transformer model.
For scale ablations, we additionally evaluate Gemma~4 E4B Instruct~\citep{gemma2026} (where the ``E'' denotes the model's \emph{effective} parameter count), a substantially smaller variant of the same model family, under identical training configurations. We include it to test whether our method is scale-agnostic and to expose training dynamics (notably policy entropy) that are harder to observe at 31B, where entropy is already low at initialization. To isolate the contributions of tool access, RL training, and inference-time scaling, we evaluate several controlled configurations of the same architecture.
The first baseline is the Gemma~4 31B Instruct model prompted with our M-Schema context and few-shot exemplars but given no tool access and no RL training. This measures the base model's zero-shot SQL ability under our evaluation harness.
The second baseline augments the same model with the full tool suite as discussed in Section~\ref{sec:tools}. The system prompt describes all three tools and the verification-first protocol but still uses no RL-trained weights, therefore, any performance gain reflects the model's pre-existing instruction-following capacity to call tools when prompted.
Comparing these two baselines quantifies how much of the final improvement is attributable to tool availability versus learned tool proficiency.

% % \paragraph{Baselines.}
% We compare against:
% \begin{itemize}[nosep,leftmargin=*]
%   \item The \emph{base} Gemma~4 Instruct model with and without tool access;
%   \item Intermediate RL checkpoints to trace training progression.
% \end{itemize}

% ──────────────────────────────────────────────────────────────────
\subsection{Training Infrastructure}
\label{sec:infra}

Training runs on a single node of 8~NVIDIA H200 GPUs.
Six GPUs are dedicated to the policy model with DeepSpeed ZeRO-3
%(stage-3 parameter sharding, CPU optimizer offload, gradient clipping at 1.0)
~\citep{rajbhandari2019zero}.
A separate vLLM server on the 2~remaining H200 GPUs (tensor parallelism${}=2$, \texttt{gpu\_memory\_utilization}${}=0.93$) handles rollout generation with a 24{,}576-token context window.
Weight synchronization between the training policy and the vLLM server occurs after each optimizer step via a dedicated NCCL group.

Key hyperparameters: learning rate $10^{-6}$ with AdamW, warmup ratio 0.03, per-device batch size~2, gradient accumulation 16~steps (effective batch: $2 \times 16 \times 6 = 192$ completions per optimizer step), sampling temperature~1.2, nucleus sampling $p = 0.95$, gradient clipping at $1.0$, one training epoch.
% !TEX root = ../main.tex
\section{Results}
\label{sec:results}

% ──────────────────────────────────────────────────────────────────
\subsection{Main Results}

Table~\ref{tab:main-results} presents the main results of \modelname on the BIRD-SQL development benchmark.
We isolate the contribution of each training stage
%under a common evaluation protocol
: the supervised warm-start (SFT), agentic RL from the base model (RFT), and the compound SFT$\to$RFT pipeline.
%All configurations use the full tool suite;
The compound SFT$\to$RFT model is our strongest system, reaching \textbf{74.32\%~EX} under single-pass decoding and \textbf{74.77\%~EX} with execution-based self-consistency.

\begin{table}[t]
\centering\small
\caption{Training-stage ablation on BIRD-SQL dev (1,534 questions) under a common evaluation protocol (parallel execution, 30s timeout, set-based comparison). All rows use the tool suite. SFT = supervised fine-tuning on rejection-sampled privileged-teacher and self-traces (Section~\ref{sec:sft}); RFT = agentic reinforcement fine-tuning via GRPO from the base model (Section~\ref{sec:grpo}); SC = execution-based self-consistency (majority vote over 16 samples). $\Delta$ is temp-0 (greedy) EX relative to the base model.}
\label{tab:main-results}
\begin{tabular}{lcrrr}
\toprule
\textbf{Configuration} & \textbf{SC} & \textbf{$\Delta$ vs.\ Base} & \textbf{EX (\%)} & \textbf{pass@16 (\%)} \\
\midrule
Gemma~4 31B Instruct (base)              & \ding{55} & ---     & 71.19 & 76.9 \\
\midrule
+ SFT                           & \ding{55} & +1.50   & 72.69 & \textbf{81.29} \\
+ RFT (from base)               & \ding{55} & +2.47   & 73.66 & 77.2 \\
+ RFT + SC                      & \ding{51} & +2.93   & 74.12 & --- \\
\midrule
\textbf{+ SFT $\to$ RFT}        & \ding{55} & +3.13   & \textbf{74.32} & \textbf{81.94} \\
\textbf{+ SFT $\to$ RFT + SC}   & \ding{51} & +3.58   & \textbf{74.77} & --- \\
\bottomrule
\end{tabular}
\end{table}

% The progression tells a clear story, and the two stages act on \emph{different} axes of performance.

\paragraph{SFT alone.}
The supervised warm-start (Section~\ref{sec:sft}) lifts temp-0 accuracy to 72.69\% (+1.50\pp) and raises pass@16 coverage from 76.9\% to \textbf{81.29\%}.
The disproportionate pass@16 gain is the intended effect of the pipeline: the privileged-teacher traces inject verified solutions for hard questions the base policy could not reach at any sampling budget. Thus, broadening the pool of questions the model can solve with at least one correct query among its 16 samples.

\paragraph{Agentic RL alone.}
Reinforcement fine-tuning from the base model reaches 73.66\% (74.12\% with self-consistency).
RL improves temp-0 accuracy more than SFT (2.47\pp\ vs.\ 1.50\pp) but leaves pass@16 coverage essentially unchanged (77.2\%). This shows that RL \emph{sharpens} the policy toward programs it can already sometimes find, teaching it \emph{when} to verify a draft query, \emph{what} evidence to retrieve, and \emph{how} to revise faulty SQL, rather than expanding coverage of previously unsolved questions.

\paragraph{SFT$\to$RFT (compound pipeline).}
Initializing RL from the SFT checkpoint combines both effects and yields our best model giving EX of \textbf{74.32\%} temp-0 and \textbf{74.77\%} with self-consistency, with pass@16 held at 81.94\%.
Relative to RL from the base model this is +0.66\pp\ temp-0 and +0.65\pp\ self-consistency at parity-or-better pass@16.
Crucially, agentic RL preserves the SFT-expanded coverage while converting it into higher single-pass accuracy. The warm-start does not merely provide an easier optimization start but raises the reachable temp-0 ceiling.
This complementarity (SFT widens capability on hard examples; RL hones single-pass reliability) mirrors the supervised-then-reinforced findings of Arctic-Text2SQL-R1~\citep{yao2026arctic}.

% ──────────────────────────────────────────────────────────────────
\subsection{BIRD Dev Leaderboard Comparison}

Table~\ref{tab:leaderboard} positions \modelname against the top entries on the BIRD single-model development leaderboard (public entries as of Aug26~2026).

\begin{table}[t]
\centering\small
\caption{BIRD-SQL dev leaderboard (single-model track, 1,534 questions, Aug~26~2026). SC indicates self-consistency. Shaded rows show our experiments and contributions. $^\ddagger$pass@16 majority vote.}
\label{tab:leaderboard}
\setlength{\tabcolsep}{4pt}
\begin{tabular}{r@{\hspace{6pt}}l@{\hspace{8pt}}r@{\hspace{6pt}}l@{\hspace{8pt}}r}
\toprule
\textbf{Rank} & \textbf{System} & \textbf{Params} & \textbf{SC} & \textbf{EX (\%)} \\
\midrule
\rowcolor{blue!8}
1 & \textbf{ReToolSQL (SFT$\to$RFT) + SC$^\ddagger$} & 31B & Many & \textbf{74.77} \\
\rowcolor{blue!8}
2 & \textbf{ReToolSQL (SFT$\to$RFT)} & 31B & ---  & \textbf{74.32} \\
3 & Gemini-SQL2 & UNK & Many & 74.12 \\
\rowcolor{blue!8}
3 & \textbf{ReToolSQL (RFT) + SC$^\ddagger$} & 31B & Many & \textbf{74.12} \\
4 & SiriusAI-Text2SQL (Tencent)              & 27B & Many & 73.70 \\
\rowcolor{blue!8}
5 & \textbf{ReToolSQL (RFT)}                  & 31B & ---  & \textbf{73.66} \\
\addlinespace[2pt]
6 & Gemini-SQL (Google)                         & UNK & Many & 73.27 \\
7 & Q-SQL (AWS)                                 & 30B-3B-MoE & Many & 72.99 \\
%7 & SiriusAI-Text2SQL-v2 (Tencent)              & 32B & Many & 72.82 \\
8 & Sophon-Text2SQL (ByteDance)                 & 32B & Many & 72.43 \\
9 & Arctic-Text2SQL-R1 (Snowflake)              & 32B & Few  & 72.20 \\
10 & Spektr-SQL (Amazon Ads)                     & 30B-3B-MoE & Few  & 72.10 \\
\addlinespace[2pt]
\rowcolor{blue!8}
11 & \textbf{Gemma~4 31B Instruct + Tool (no RL)}   & 31B & ---  & \textbf{71.71} \\
11 & ShepherdSQL (JD)                            & 30B-3B-MoE & Many & 71.71 \\
12 & Arctic-Text2SQL-R1-14B                      & 14B & Few  & 71.40 \\
13 & FTEL-Text2SQL (FPT)                         & 30B-3B-MoE & Few  & 71.19 \\
14 & Jiayin-Pangu-14B                            & 14B & Many & 71.10 \\
15 & Databricks RLVR 32B                         & 32B & ---  & 70.80 \\
16 & Arctic-Text2SQL-R1-7B                       & 7B  & Few  & 70.70 \\
\addlinespace[2pt]
\rowcolor{blue!8}
24 & \textbf{Gemma~4 31B Instruct baseline}         & 31B & ---  & \textbf{69.69} \\
\bottomrule
\end{tabular}
\\[0.3em]
\end{table}

In our experiments, the compound SFT$\to$RFT model achieves \textbf{74.77\%} with self-consistency, ahead of Gemini-SQL2 (74.12\%), Gemini-SQL (Google, 73.27\%), Q-SQL (AWS, 72.99\%), and Arctic-Text2SQL-R1 (Snowflake, 72.20\%), all of which rely on multi-sample voting or MoE architectures.
Even single-pass, the SFT$\to$RFT model (74.32\%) and RFT-only with self-consistency (74.12\%) are competitive with the strongest published entries.
Notably, \modelname attains the top position with a single dense 31B model and simple execution-based self-consistency, without an external retriever, a learned selector, or a multi-component pipeline.
% !TEX root = ../main.tex
\section{Analysis}
\label{sec:analysis}

% ──────────────────────────────────────────────────────────────────
\subsection{Component Isolation}

To separate the contributions of RL training and tool access, we factor the two dimensions in Table~\ref{tab:ablation}. Each row represents the training condition (no RL vs.\ agentic RL) and each column represents the inference condition (no tools vs.\ tools available).

\begin{table}[t]
\centering\small
\caption{Component isolation: RL training $\times$ tool access on BIRD-SQL dev (1,534 questions). The interaction effect (+1.5\pp\ beyond additive components) confirms that agentic RL specifically teaches the model to leverage tools effectively.}
\label{tab:ablation}
\setlength{\tabcolsep}{6pt}
\begin{tabular}{l cc c}
\toprule
& \multicolumn{2}{c}{\textbf{Inference-Time Tools}} & \\
\cmidrule(lr){2-3}
\textbf{Training} & \textbf{No Tools} & \textbf{+ Tools} & \textbf{Tool $\Delta$} \\
\midrule
%\rowcolor{gray!6}
No RL (Gemma~4 31B Instruct) & 69.69\% & 71.71\% & \textcolor{gray}{+2.02} \\
Agentic RL (\modelname) & 70.14\% & \textbf{73.66\%} & \textbf{+3.52} \\
\addlinespace[3pt]
\midrule
\textbf{RL $\Delta$}    & \textcolor{gray}{+0.45} & \textbf{+1.95} & \\
\midrule
%\rowcolor{blue!8}
\textbf{+ Self-Consistency (SC@16)} & --- & \textbf{74.12\%} & \textbf{+4.43 vs.\ base} \\
\bottomrule
\end{tabular}
\end{table}

The $2 \times 2$ factoring reveals a clear interaction effect. Tool access alone (without RL) improves the base model by a modest 2.02\pp, while RL training without tools yields only 0.45\pp. 
% indicating that the RL-trained model has developed latent capabilities that remain dormant until tools are available.
When both the factors are present, the gain is 3.97\pp\ over the base, which exceeds the sum of individual contributions
%(2.02 + 0.45 = 2.47)
by 1.50\pp.
Viewed from the tool-calling column, RL training nearly doubles the benefit of tool access (3.52\pp\ vs.\ 2.02\pp), confirming that agentic RL specifically teaches the model on \emph{how} to leverage execution feedback rather than merely making tools available.

% ──────────────────────────────────────────────────────────────────
\subsection{Pass@k and Sampling Analysis}
\label{sec:passk}

We analyze the relationship between sampling budget and accuracy using the pass@$k$ metric (fraction of questions where at least one of $k$ samples is correct).

For \modelname,
%(31B with tools),
pass@1 (greedy) achieves 73.66\%, self-consistency over 16~samples reaches 74.12\%, and the pass@16 upper bound (the fraction of questions for which at least one of the 16 sampled trajectories is execution-correct, i.e., the accuracy attainable by a perfect candidate selector) is approximately 77.2\%.
The gap between self-consistency and this upper bound is 3.08\pp ~which reveals a substantial unrealized potential. This showcases that on these questions at least one of the 16~samples produces the correct SQL, but majority voting selects a wrong cluster.
This suggests that improved selection mechanisms, such as learned verifiers or reward-model reranking, could unlock further gains without additional training. This will be addressed in our future work.

The modest self-consistency improvement (+0.46\pp) is explained by the model's high per-sample consistency. The 31B policy produces only 3.37~unique SQL strings per 16~candidates on average. When the model can solve a query, the correct answer already dominates the sample set and when it cannot most candidates cluster into a single incorrect formulation.
This behavior is different than the smaller models (Section~\ref{sec:scale}), which exhibit greater diversity and correspondingly larger self-consistency gains. Although, the overall accuracy still remains lesser than the larger capacity model.

% ──────────────────────────────────────────────────────────────────
\subsection{Ablation: Model Scale}
\label{sec:scale}

To showcase that the presented idea is model and scale agnostic, we replicate the full training setup on Gemma~4 E4B Instruct (\cite{gemma2026}) using identical reward functions, tool access, DAPO dynamic sampling, and beta scheduling to study scale effects (Table~\ref{tab:scale}).

\begin{table}[t]
\centering\small
\caption{Scale comparison: Gemma~4 E4B Instruct vs.\ Gemma~4 31B Instruct. SC = self-consistency (pass@16).}
\label{tab:scale}
\begin{tabular}{lrrrr}
\toprule
\textbf{Model} & \textbf{Base EX} & \textbf{RL EX} & \textbf{$\Delta$RL} & \textbf{SC EX} \\
\midrule
Gemma~4 E4B  & 65.12\% & 68.45\% & +3.33\pp & 70.73\% \\
Gemma~4 (31B)     & 71.71\% & 73.66\% & +1.95\pp & 74.12\% \\
\bottomrule
\end{tabular}
\\[0.3em]
{\footnotesize Base EX is with tool access; $\Delta$RL is the gain from agentic RL training.}
\end{table}

Under greedy decoding, both model sizes improve after agentic RL training. Gemma~4 E4B Instruct rises from 65.12\% to 68.45\%, a gain of 3.33\pp, while the 31B model rises from 71.71\% to 73.66\%, a gain of 1.95\pp, both with tool access. The smaller model shows a somewhat larger marginal gain. One likely explanation is that the 31B model already starts from a stronger base policy, leaving less room for RL to further improve execution accuracy. We see a similar pattern under inference-time sampling. E4B produces a broader set of candidates, with 6.31 unique SQL strings per 16 samples compared with 3.37 for the 31B model, and this gives it more room to benefit from aggregation. Its pass@$k$ gain from pass@$1$ to pass@$16$ is 16.15\pp, compared with 6.28\pp for the 31B model. Taken together, these results show that the smaller model improves under both RL training and inference-time aggregation, although the 31B model remains stronger in absolute terms and is therefore used as the deployed configuration.

\subsection{Ablation: Entropy Preservation via Beta Scheduling}
\label{sec:beta}

We compare two Gemma~4 E4B Instruct training configurations to study the effect of KL regularization on learning dynamics:
\begin{enumerate}[nosep,leftmargin=*]
  \item No KL penalty ($\beta = 0$ throughout training).
  \item Decaying beta schedule: $\beta = 0.005$ for steps 0--40, $\beta = 0.001$ for steps 40--80, and $\beta = 0$ from step~80 onward.
\end{enumerate}

The beta schedule yields 2.09\pp\ better greedy accuracy and 1.96\pp\ better self-consistency accuracy. Critically, it preserves entropy approximately $2\times$ longer, as we see from Figure~\ref{fig:entropy-beta}, where the entropy declines from 0.40 to 0.20 under the schedule, versus 0.40 to 0.10 without it with $\beta = 0$ throughout training as shown in Figure~\ref{fig:entropy-collapse}. This is consistent with the observation that identifies policy entropy as a central determinant of sustained improvement in RL for reasoning models~\citep{cui2025entropy}. Also, it can be seen that in the No KL penalty run, entropy collapses within 40 training steps and accuracy stalls, whereas, in the decaying beta-schedule run slower entropy decline permits sustained learning past step~100.

At 31B scale, entropy is already low at initialization (0.17 vs.\ 0.40 for E4B) and declines further during training (reaching 0.02--0.08). The beta schedule still produces the best trajectory, monotonic improvement through training, but the limited starting entropy partly explains why RL gains are smaller at 31B.
%: the model has limited exploration capacity to discover substantially better SQL policies.

% ──────────────────────────────────────────────────────────────────
\subsection{Training Dynamics}
\label{sec:dynamics}

\paragraph{DAPO Dynamic Sampling.}
The trainer uses oversampling to find heterogeneous prompt groups (groups where not all 16 samples produce the same reward).
At 31B scale with 192 groups attempted per step, typically only 1--15 groups exhibit reward variance, yielding fill rates of 8--67\%.
This reflects the model's high per-sample consistency: when it can solve a query, all 16 samples tend to be correct; when it cannot, most fail identically.
E4B exhibits higher heterogeneity, enabling the policy gradient to receive signal from more diverse training groups per step.

% \paragraph{Checkpoint Progression.}
% The 31B training trajectory is non-monotonic when evaluated with tool calling on the full dev set: early checkpoints achieve 71.71\% and the final model reaches 73.66\%, but intermediate checkpoints exhibit substantial variance.
% Internal training-time evaluations (on a subset with a shorter timeout) show oscillations with intermittent gradient norm spikes ($7$--$20\times$) followed by consolidation phases.
% This pattern suggests that the model periodically discovers and integrates new SQL generation strategies, with temporary disruption before stabilizing at higher accuracy.

% ──────────────────────────────────────────────────────────────────
\subsection{Difficulty-Stratified Analysis}

\begin{table}[t]
\centering\small
\caption{Accuracy by difficulty level for key configurations. All rows use the same difficulty annotation (860/443/231 split).}
\label{tab:difficulty}
\begin{tabular}{lrrr}
\toprule
\textbf{Configuration} & \textbf{Simple} & \textbf{Moderate} & \textbf{Challenging} \\
\midrule
Gemma~4 31B Instruct (no tools)     & 69.30\% & 67.72\% & 74.89\% \\
\modelname (no tools)     & 70.23\% & 68.85\% & 72.29\% \\
\modelname (tools)        & 73.60\% & 72.91\% & 75.32\% \\
\modelname + SC@16        & 73.95\% & 73.81\% & 75.32\% \\
\bottomrule
\end{tabular}
\end{table}

Table~\ref{tab:difficulty} reveals that RL training progressively improves Moderate and Challenging tiers.
The base model without tools shows a striking pattern where Challenging accuracy (74.89\%) exceeds Simple (69.30\%), possibly because challenging questions have richer hints that aid the model.
After agentic RL training with tools, \modelname achieves more balanced performance across all tiers
%(73.60\%/72.91\%/75.32\%),
with substantial gains on Moderate (5.19\pp\ vs.\ base) indicating that GRPO teaches the model when tools are helpful versus when direct reasoning suffices.
The Challenging tier maintains the highest absolute accuracy (75.32\%) for the RL model, consistent with tool-augmented verification being most valuable for complex multi-step queries.
% !TEX root = ../main.tex
\section{Conclusion}
\label{sec:conclusion}

We present \modelname, a text-to-SQL system that trains a single dense 31B language model using tool-augmented reinforcement learning. During training, the model interacts directly with database tools inside the RL environment, including a sandboxed SQL executor, a column profiler, and a BM25-based value searcher. By rewarding effective tool use, the model learns a verification-first policy where it drafts SQL queries, executes them, diagnoses errors or mismatches, and repairs the query within a unified propose–verify–repair loop.
% We present \modelname, a text-to-SQL system that trains a single dense 31B language model to solve database questions through tool-augmented reinforcement learning.
% By placing database tools, a sandboxed SQL executor, a column profiler, and a BM25 value searcher, directly inside the reinforcement-learning environment and rewarding their effective use during training, the model learns a verification-first policy that drafts, executes, diagnoses, and repairs SQL within a unified propose--verify--repair cycle.
On the BIRD-SQL development benchmark, agentic RL from the base model achieves 73.66\% execution accuracy under single-pass decoding and 74.12\% with execution-based self-consistency. Further adding a supervised warm-start on rejection-sampled reasoning traces (the SFT$\to$RFT pipeline) raises this to \textbf{74.32\%} single-pass and \textbf{74.77\%} with self-consistency (our strongest configuration), showing that agentic reinforcement learning can yield strong performance while preserving a relatively simple serving architecture.

% Our results validate a core insight: \emph{treating SQL generation as a verification problem} rather than a one-shot generation task is the key to robust text-to-SQL performance.

More broadly, the paper highlights the following practical lessons:
\begin{enumerate}[nosep,leftmargin=*]
  \item \textbf{Robust text-to-SQL depends on training objectives that explicitly teach models to use execution feedback for iterative correction.} Tool access alone provides modest gains (+2\pp) and with RL training over tool-use trajectories provides substantially more (+4\pp).
  % \item \textbf{Competitive performance need not require heavyweight multi-component pipelines.} A single-policy system can remain competitive when trained to verify and repair its own outputs.
  \item \textbf{Entropy preservation is critical for RL at scale.} Beta scheduling maintains exploration capacity, producing sustained learning even when the base model is already strong.
  \item \textbf{Supervised warm-start and RL act on complementary axes.} A supervised stage on execution-verified privileged-teacher traces expands pass@$k$ coverage of hard questions, while agentic RL sharpens single-pass accuracy; composing them (SFT$\to$RFT) outperforms either stage alone.
  % \item \textbf{Training for iterative correction matters more than tool access.} Simply making tools available yields a modest +2\pp; RL over multi-turn trajectories yields +4\pp. The gap shows that tool proficiency must be learned, not just prompted.
  % \item \textbf{Heavyweight pipelines are not required.} A single model that can verify and repair its own outputs is competitive with multi-component systems costing far more at inference.
  % \item \textbf{Entropy preservation is critical.} Without beta scheduling, the 31B policy collapses to near-deterministic output within 40 steps and stops improving.
  % \item \textbf{SFT and RL fix different problems.} SFT on privileged traces widens pass@$k$ coverage; RL sharpens single-pass accuracy. Neither subsumes the other.
\end{enumerate}
% \end{enumerate}

% \paragraph{Limitations and Future Work.}
% Our results are currently limited to the BIRD-SQL development set; official test-set submission is planned as future work.
% The self-consistency gap between majority vote (74.12\%) and oracle pass@16 (${\approx}$79.9\%) suggests that improved selection mechanisms---potentially learned selectors trained on execution features---could recover additional accuracy.
% Future work should study transfer beyond SQLite, extend the framework to broader enterprise database settings with larger schemas and proprietary data, and explore richer reward formulations that capture semantic partial credit, efficiency, and robustness under evolving schemas.

From an architectural standpoint, \modelname demonstrates that a single-policy system can compete with multi-component pipelines.
The model serves both generation and verification without an external retriever, a learned selector, or mixture-of-experts routing.
This simplicity makes the system suitable for deployment on commodity hardware, with inference requiring only a single GPU for the 31B model under standard serving frameworks.

\subsubsection*{Acknowledgements}
We thank Vijay Parthasarathy, Chul Lee, Rajaram Kumar, Yateen Dhareshwar, Bharath Subramanian, Venkat Vemuri, Xiaonan Wang, Ganapathy Mani, Soham Palande, and other members of the JPMC AI Technology team for their collaboration.

% ── References ────────────────────────────────────────────────────
\bibliographystyle{plainnat}
\bibliography{references}

% ── Appendix ──────────────────────────────────────────────────────
\clearpage
\begin{appendices}

\section{Worked Example: Multi-Turn Tool Trajectory}
\label{app:example}

Figure~\ref{lst:trajectory} shows a representative trajectory on the \texttt{california\_schools} database.
Each tool call ends the turn and is answered by an environment observation before the next turn begins.
The trajectory illustrates the three core capabilities learned through RL: (1)~execution-triggered error detection, (2)~targeted evidence retrieval, and (3)~structured repair with safety guards.

\begin{figure}[H]
\definecolor{turncolor}{HTML}{2B579A}
\definecolor{reasoncolor}{HTML}{5B5B5B}
\definecolor{toolcolor}{HTML}{1A6B3F}
\definecolor{obscolor}{HTML}{8B4513}
\definecolor{finalcolor}{HTML}{1A3A6B}
\newcommand{\turnhdr}[2]{\smallskip\noindent{\color{turncolor}\sffamily\bfseries Turn~#1}\enspace{\small\color{turncolor!70!black}(#2)}\par\smallskip}
\newcommand{\reason}[1]{{\small\color{reasoncolor}\itshape\hangindent=1em #1}\par}
\newcommand{\toolcall}[1]{{\small\ttfamily\color{toolcolor} #1}\par}
\newcommand{\obs}[1]{\hfill\begin{minipage}[t]{0.92\linewidth}{\footnotesize\ttfamily\color{obscolor}$\boldsymbol{\rightarrow}$~#1}\end{minipage}\par}
\begin{tcolorbox}[
  enhanced,
  breakable,
  title={\small Multi-turn agentic trajectory on \texttt{california\_schools} (abbreviated)},
  fonttitle=\sffamily\bfseries,
  coltitle=white,
  colbacktitle=turncolor!85!black,
  colback=gray!3,
  colframe=turncolor!40!black,
  boxrule=0.5pt,
  arc=2pt,
  left=6pt, right=6pt, top=4pt, bottom=4pt,
  before upper={\small}
]

\turnhdr{1}{draft + execute}
\reason{ExpectedOutputColumns = [max\_rate]. CandidateSQL drafts the ratio; verify execution first.}
\toolcall{call: sqlite\_query\{db\_id: california\_schools, sql: SELECT MAX(CAST(f."Free Meal Count (K-12)" AS REAL) / f."Enrollment (K-12)") FROM frpm f JOIN schools s ON f.CDSCode=s.CDSCode WHERE f."County Name"='alameda' AND s.Charter=1\}}
\obs{\{"columns":["max\_rate"], "rows":[], "row\_count":0\} \textnormal{\color{red!70!black}\bfseries --- empty!}}

\tcbline

\turnhdr{2}{repair the literal}
\reason{Zero rows returned; the county literal may not match stored values.}
\toolcall{call: bm25\_search\_sqlite\{db\_id: california\_schools, table: frpm, column: "County Name", query: alameda, top\_k: 5\}}
\obs{[\{"value":"Alameda", "bm25":8.71\}, \{"value":"Alameda County", "bm25":5.2\}]}

\tcbline

\turnhdr{3}{check denominator scale/nulls}
\reason{Guard division: inspect Enrollment column for nulls and zeros.}
\toolcall{call: sqlite\_peek\{db\_id: california\_schools, table: frpm, columns: ["Enrollment (K-12)"], limit: 20\}}
\obs{\{"Enrollment (K-12)": \{"type":"REAL", "nulls":1037, "min":1.0, "max":6800.0\}\}}

\tcbline

\turnhdr{4}{finalize}
\reason{Use `Alameda'; cast to REAL; require Enrollment > 0.}

\smallskip
\begin{tcolorbox}[
  enhanced, boxrule=0.4pt, colframe=finalcolor!60, colback=finalcolor!4,
  arc=1.5pt, left=4pt, right=4pt, top=2pt, bottom=2pt,
  fontupper=\small\ttfamily
]
\color{finalcolor}%
\textbf{<final\_answer>}\\
SELECT MAX(CAST(f."Free Meal Count (K-12)" AS REAL)/f."Enrollment (K-12)")\\
FROM frpm f JOIN schools s ON f.CDSCode=s.CDSCode\\
WHERE f."County Name"='Alameda' AND s.Charter=1 AND f."Enrollment (K-12)">0;\\
\textbf{</final\_answer>}
\end{tcolorbox}
\end{tcolorbox}
\captionof{figure}{A representative \modelname trajectory. The policy drafts SQL, detects an empty result via execution, retrieves the correct literal with BM25 search, profiles the denominator column to discover nulls, and emits a guarded final query.}
\label{lst:trajectory}
\end{figure}

% ──────────────────────────────────────────────────────────────────
\section{Execution-Based Self-Consistency}
\label{app:self-consistency}

We use execution-based self-consistency as a training-free inference procedure.
For a question, schema, and tool set, the policy samples 16 candidate trajectories at temperature~1.2 and extracts the final SQL from each.
Each candidate is validated under the same safety constraints used during training, executed against the target database, and mapped to its result set.
Candidates that fail to execute or return an empty result are discarded.
The remaining candidates are clustered by result-set equality, and the system returns the representative SQL from the largest cluster, breaking ties by earliest sample index and then by shorter SQL length.

This procedure improves robustness by favoring semantically stable answers over isolated generations.
Because semantically distinct hallucinations tend to produce unique result sets that scatter across singleton clusters, while correct formulations converge on a shared output, the majority-vote mechanism has an inherent bias toward correctness.

% ──────────────────────────────────────────────────────────────────
\section{Tool Definitions}
\label{app:tools}

This appendix specifies the concrete API contract for each tool: full call signatures, parameter names and defaults, the resource limits enforced by the sandbox, and the exact structure returned to the policy (Table~\ref{tab:tool-api}).

\begin{center}
\footnotesize
\setlength{\tabcolsep}{8pt}
\renewcommand{\arraystretch}{1.4}
\captionof{table}{Read-only database tool API (complementing the conceptual descriptions in Section~\ref{sec:tools}). All three tools enforce read-only access, preventing schema modifications or data writes during both training and inference.}
\label{tab:tool-api}
\begin{tabular}{@{}>{\raggedright\arraybackslash}p{3.9cm} >{\raggedright\arraybackslash}p{4.0cm} >{\raggedright\arraybackslash}p{6.1cm}@{}}
\toprule
\textbf{Tool \& signature} & \textbf{Enforced limits} & \textbf{Returns} \\
\midrule
\texttt{sqlite\_query} \newline \texttt{(db\_id, sql,} \newline \texttt{max\_return\_rows=100)}
  & \texttt{SELECT}/\texttt{WITH} only; VM step limit 15M; busy timeout 5\,s; row cap
  & Columns, rows, row count, and truncation flag; or a structured error message \\
\addlinespace[3pt]
\texttt{sqlite\_peek} \newline \texttt{(db\_id, table,} \newline \texttt{columns, limit=10,} \newline \texttt{where=None)}
  & Profile scan limited to 5{,}000 rows
  & Per column: declared type, null/distinct counts, value ranges, character-class summaries, top-$k$ values with frequencies, and representative samples \\
\addlinespace[3pt]
\texttt{bm25\_search\_sqlite} \newline \texttt{(db\_id, table, column,} \newline \texttt{query, top\_k=10,} \newline \texttt{where=None)}
  & Read-only; case-insensitive by default
  & $2 \times \text{top\_k}$ results (BM25-ranked plus regex-matched) with scores \\
\bottomrule
\end{tabular}
\end{center}
% ──────────────────────────────────────────────────────────────────
\section{Entropy Dynamics without KL Regularization (Gemma E4B model)}
\label{app:entropy}

\begin{figure}[h]
\centering
\includegraphics[width=\linewidth]{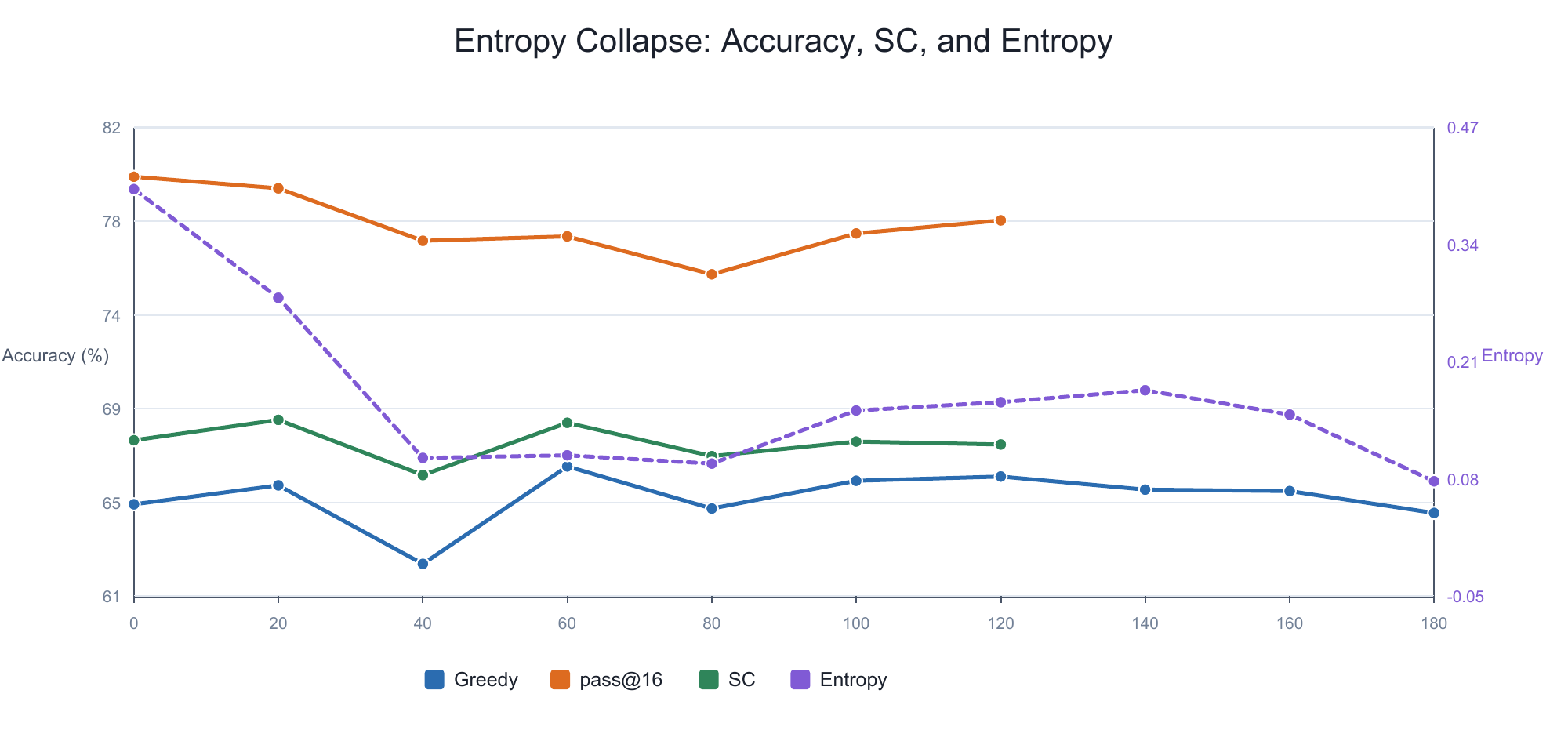}
\caption{Entropy collapse in the Gemma~4 E4B no-KL run ($\beta = 0$): execution accuracy, self-consistency (SC@16), and policy entropy versus training checkpoint (shared scale). Entropy collapses within ${\sim}40$ steps, after which accuracy plateaus while pass@$k$ headroom shrinks.}
\label{fig:entropy-collapse}
\end{figure}

Figure~\ref{fig:entropy-collapse} traces execution accuracy, self-consistency (SC@16), and policy entropy across training checkpoints for the Gemma~4 E4B run trained \emph{without} a KL penalty ($\beta = 0$).
Policy entropy falls sharply from approximately $0.40$ to $0.10$ within the first $40$ steps and remains low thereafter.
Once entropy collapses, greedy and self-consistency accuracy plateau, and pass@$k$ headroom is in fact largest at the earliest checkpoints instead of the latest.
This is precisely the failure mode the decaying beta schedule of Section~\ref{sec:beta} is designed to mitigate: retaining a small KL penalty early in training preserves exploration long enough for accuracy to keep improving.

\begin{figure}[ht]
\centering
\includegraphics[width=\linewidth]{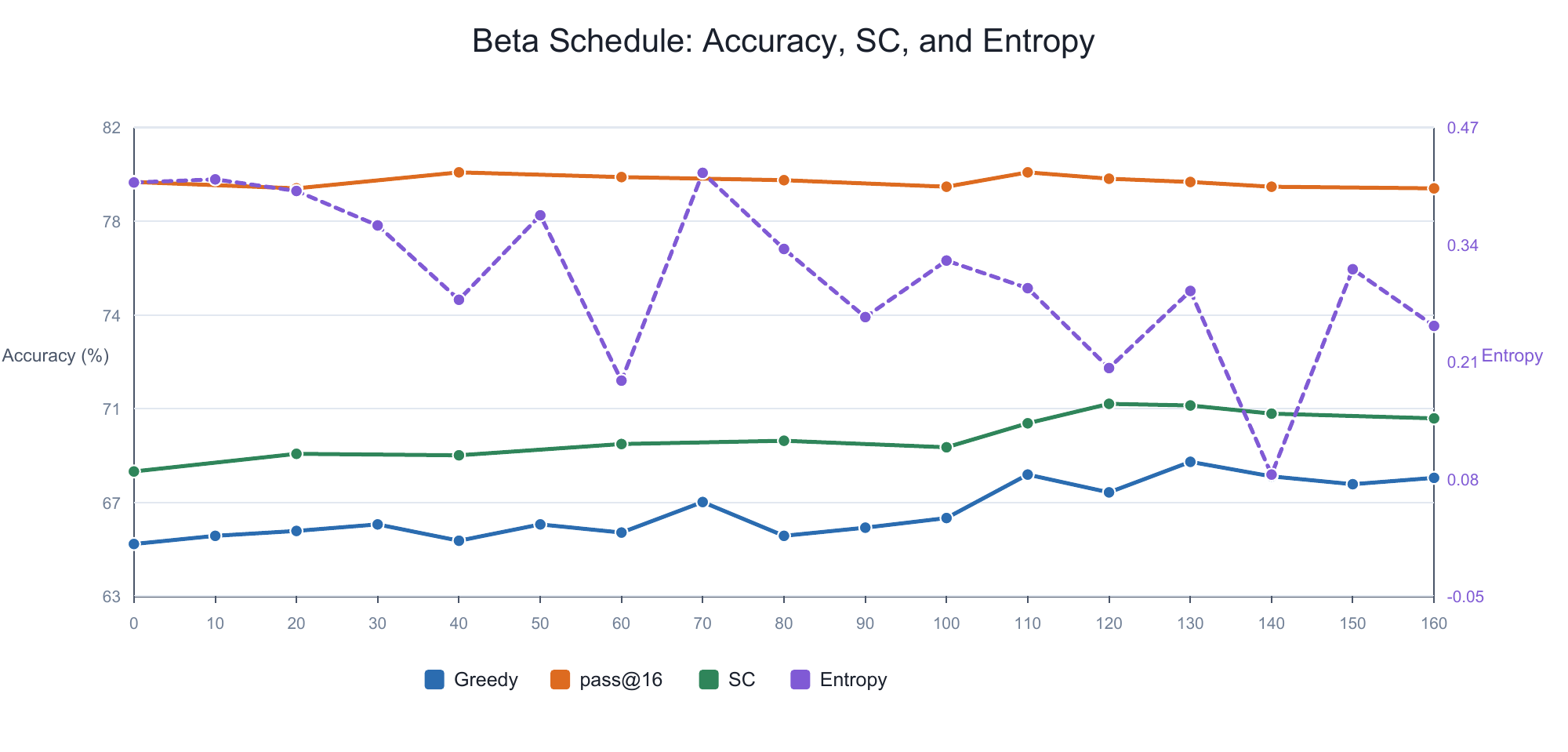}
\caption{Gemma~4 E4B trained with the decaying beta schedule ($\beta = 0.005 \to 0.001 \to 0$ at steps 0/40/80): execution accuracy, self-consistency (SC@16), and policy entropy versus training checkpoint (shared scale). Retaining a small KL penalty early preserves entropy, avoiding the collapse of Figure~\ref{fig:entropy-collapse} and sustaining learning.}
\label{fig:entropy-beta}
\end{figure}

In contrast, Figure~\ref{fig:entropy-beta} shows the same E4B configuration trained with the decaying beta schedule ($\beta = 0.005$ for steps 0--40, $\beta = 0.001$ for steps 40--80, and $\beta = 0$ from step~80 onward).
The early KL penalty slows the entropy decline and prevents the premature collapse of the no-KL run, keeping the policy exploratory; as a result, execution accuracy and self-consistency continue to improve well past the checkpoint at which the no-KL run plateaus.

\end{appendices}

\end{document}